\documentclass[11pt,a4paper]{article}
\usepackage[hyperref]{acl2020}
\usepackage{times}
\usepackage{latexsym}

\usepackage{microtype}

\usepackage{times}
\usepackage{latexsym}

\usepackage{url}

\usepackage{graphicx}  %Required
\usepackage{times}
\usepackage{latexsym}
\usepackage{amsmath}
\usepackage{amssymb}
\usepackage{enumitem}
\usepackage{multirow}
\usepackage{array}
\usepackage{booktabs}
\usepackage{multirow}
\usepackage{todonotes}
\usepackage{mathrsfs}
\usepackage{galois}
\usepackage{bm}
\usepackage{overpic}
\usepackage{algorithm}  
\usepackage{algpseudocode}
\usepackage{subfigure}
\usepackage{graphicx}
\usepackage{url}
\usepackage{hyperref}

\aclfinalcopy

\title{SentiBERT: A Transferable Transformer-Based Architecture for Compositional Sentiment Semantics}

\author{Da Yin$^\clubsuit$, Tao Meng$^\spadesuit$, Kai-Wei Chang$^\spadesuit$ \\
$^\clubsuit$ Peking University \\
$^\spadesuit$ University of California, Los Angeles  \\
  {\tt wade\_yin9712@pku.edu.cn, tmeng@cs.ucla.edu, kw@kwchang.net} \\
}

\date{}

\begin{document}
\maketitle
\begin{abstract}
We propose \texttt{SentiBERT}, a variant of BERT that effectively captures compositional sentiment semantics. 
The model incorporates contextualized representation with binary constituency parse tree to capture semantic composition. Comprehensive experiments demonstrate that \texttt{SentiBERT} achieves competitive performance on phrase-level sentiment classification. We further demonstrate that the sentiment composition learned from the phrase-level annotations on SST can be transferred to other sentiment analysis tasks as well as related tasks, such as emotion classification tasks. Moreover, we conduct ablation studies and design visualization methods to understand \texttt{SentiBERT}. We show that \texttt{SentiBERT} is better than baseline approaches in capturing negation and the contrastive relation and model the compositional sentiment semantics.

\end{abstract}

\section{Introduction}

\begin{figure}[t]
\centering
  \includegraphics[width=\linewidth, trim=200 115 130 110, clip]{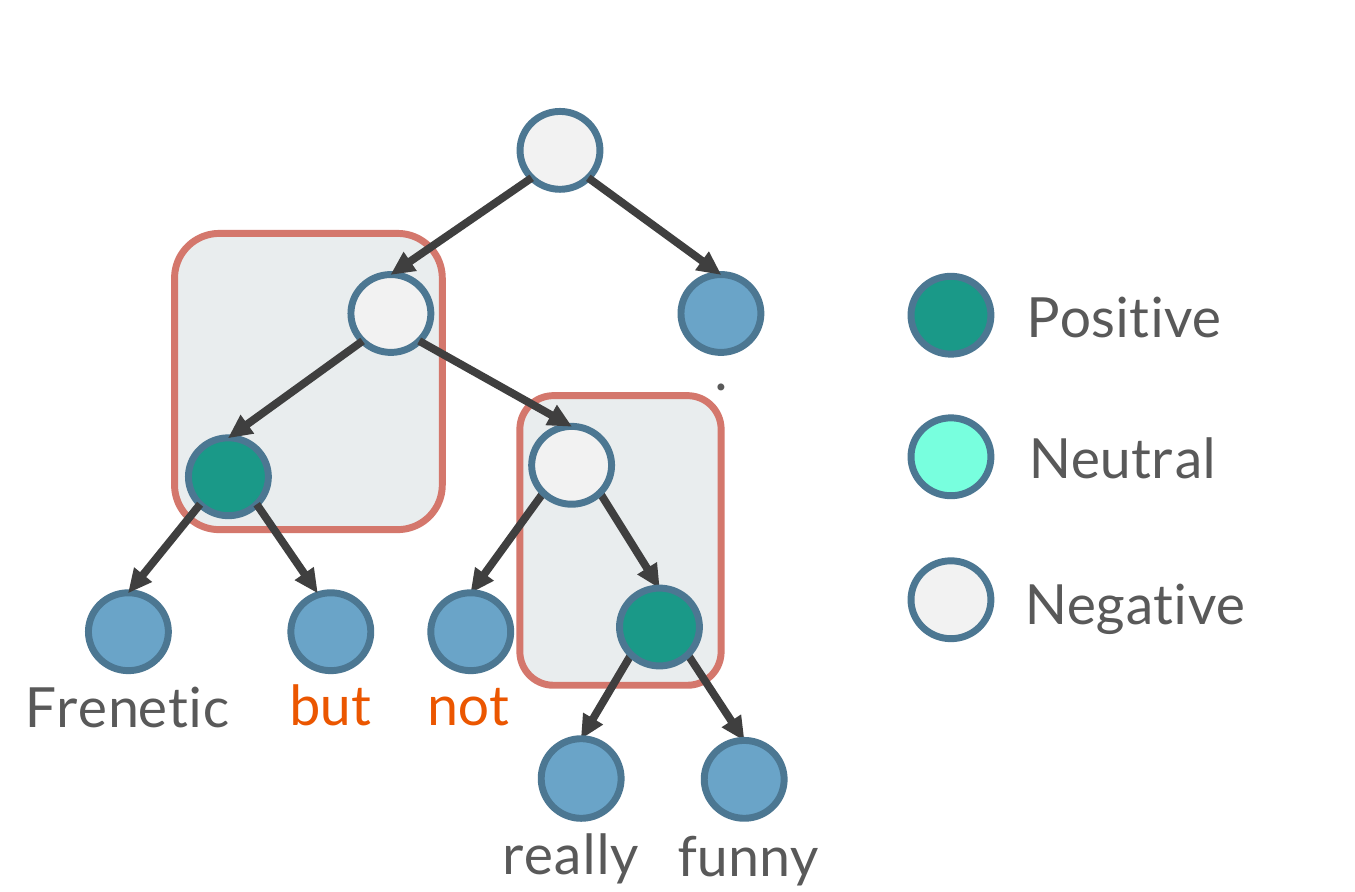}
  \caption{Illustration of the challenges of learning sentiment semantic compositionality. The blue nodes represent token nodes. The colors of phrase nodes in the binary constituency tree represent the sentiment of phrases. The red boxes show that the sentiment changes from the child node to the parent node due to negation and contrast.}
  \label{intro_exp}
\end{figure}

Sentiment analysis is an important language processing task~\cite{pang2002thumbs,pang2008opinion,liu2012sentiment}. 
One of the key challenges in sentiment analysis is to model compositional sentiment semantics. Take the sentence \textit{``Frenetic but not really funny.''} in Figure \ref{intro_exp} as an example. The two parts of the sentence are connected by \textit{``but''}, which reveals the change of sentiment. Besides, the word \textit{``not''} changes the sentiment of \textit{``really funny''}. 
These types of negation and contrast are often difficult to handle when the sentences are complex~\cite{socher2013recursive,tay2018attentive,xu2019failure}. 

In general, the sentiment of an expression is determined by the meaning of tokens and phrases and the way how they are syntactically combined.  Prior studies consider explicitly modeling compositional sentiment semantics over constituency structure with recursive neural networks \cite{socher2012semantic,socher2013recursive}. However, these models that generate representation of a parent node by aggregating the local information from child nodes, overlook the rich association in context. 

In this paper, we propose \texttt{SentiBERT} to incorporate recently developed contextualized representation models~\cite{devlin2019bert,liu2019roberta} with the recursive constituency tree structure to better capture compositional sentiment semantics. Specifically, we build a simple yet effective attention network for composing sentiment semantics on top of BERT~\cite{devlin2019bert}. During training, we follow BERT to capture contextual information by masked language modeling. In addition, we instruct the model to learn composition of meaning by predicting sentiment labels of the phrase nodes.

\begin{figure*}[!t]
\centering
  \includegraphics[width=\linewidth, trim=50 40 10 60,clip]{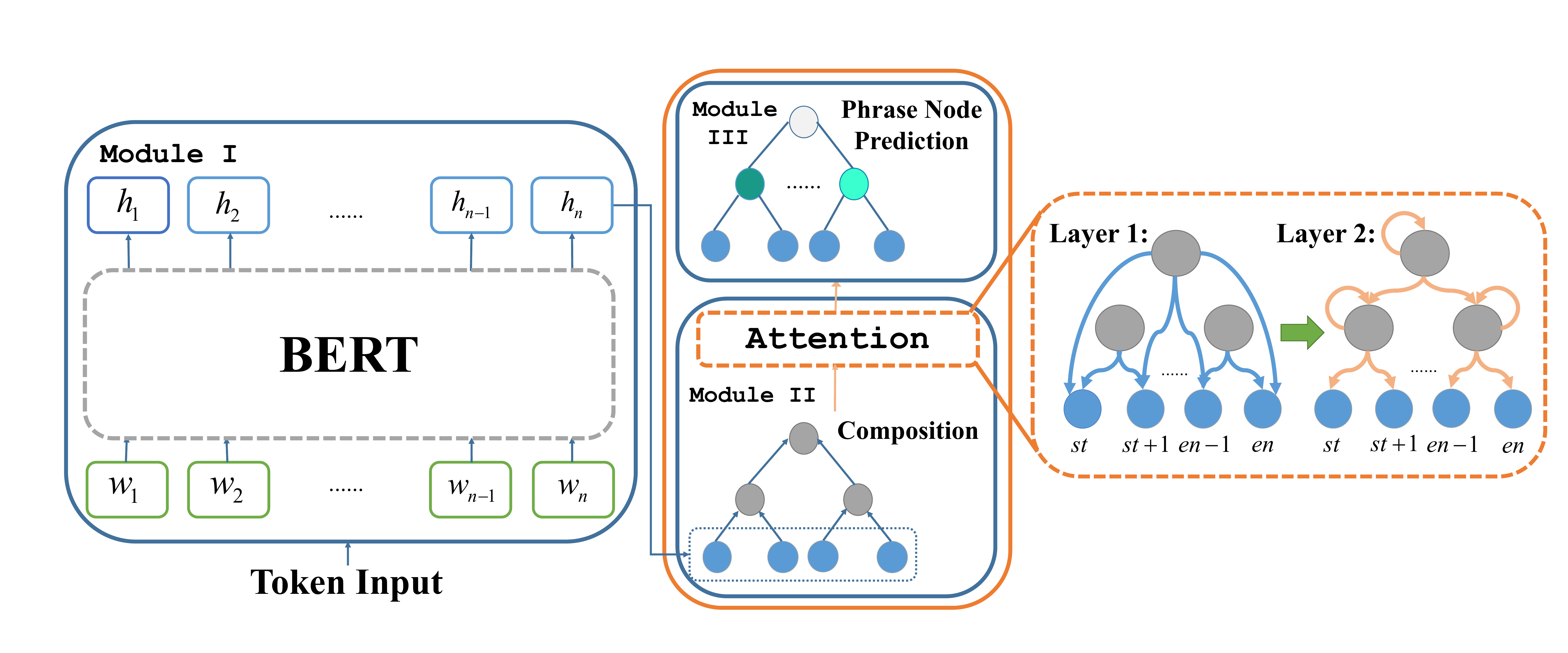}
  
  \caption{The architecture of \texttt{SentiBERT}. Module I is the BERT encoder; Module II denotes the semantic composition module based on an attention mechanism; Module III is a predictor for phrase-level sentiment. The semantic composition module is a two layer attention-based network (see Section \ref{bert}) The first layer (\textbf{Attention to Tokens}) generates representation for each phrase based on the token it covers and the second layer (\textbf{Attention to Children}) refines the phrase representation obtained from the first layer based on its children.}
  \label{sentibert}
\end{figure*}

Results on phrase-level sentiment classification on Stanford Sentiment Treebank (SST)~\cite{socher2013recursive} indicate that \texttt{SentiBERT} improves significantly over recursive networks and the baseline BERT model. As phrase-level sentiment labels are expensive to obtain, we further explore if the compositional sentiment semantics learned from one task can be transferred to others. In particular, we find that \texttt{SentiBERT} trained on SST can be transferred well to other related tasks such as twitter sentiment analysis~\cite{rosenthal2017semeval} and emotion intensity classification~\cite{mohammad2018semeval} and contextual emotion detection~\cite{chatterjee2019semeval}. 
Furthermore, we conduct comprehensive quantitative and qualitative analyses to evaluate the effectiveness of \texttt{SentiBERT} under various situations and to demonstrate the semantic compositionality captured by the model. The source code is available at \url{https://github.com/WadeYin9712/SentiBERT}.

\section{Related Work}
\paragraph{Sentiment Analysis}
Various approaches have been applied to build a sentiment classifier, including feature-based methods~\cite{hu2004mining,pang2004sentimental}, recursive neural networks~\cite{socher2012semantic,socher2013recursive,tai-etal-improved}, convolution neural networks~\cite{D14-1181} and recurrent neural networks~\cite{liu2015multi}. Recently, pre-trained language models such as ELMo~\cite{peters2018deep}, BERT~\cite{devlin2019bert} and SentiLR~\cite{ke2019sentilr} achieve high performance in sentiment analysis by constructing contextualized representation. Inspired by these prior studies, we design a transformer-based neural network model to capture compositional sentience semantics by leveraging binary constituency parse tree. 

\paragraph{Semantic Compositionality}
Semantic composition~\cite{pelletier1994principle} has been widely studied in NLP literature. For example, \citet{mitchell2008vector} introduce operations such as addition or element-wise product to model compositional semantics. The idea of modeling semantic composition is applied to various areas such as sentiment analysis~\cite{socher2013recursive,zhu-etal-2016}, semantic relatedness~\cite{marelli2014semeval} and capturing sememe knowledge~\cite{qi2019modeling}. In this paper, we demonstrate that the syntactic structure can be combined with contextualized representation such that the semantic compositionality can be better captured. Our approach resembles to a few recent attempts~\cite{harer2019tree,wang-tree} to integrate tree structures into self-attention. However, our design is specific for the semantic composition in sentiment analysis.

\section{Model}
\label{sect:details}
We introduce \texttt{SentiBERT}, a model that captures compositional sentiment semantics based on constituency structures of sentences. \texttt{SentiBERT} consists of three modules: 1) BERT; 2) a semantic composition module based on an attention network; 3) phrase and sentence sentiment predictors.
The three modules are illustrated in Figure \ref{sentibert} and we provide an overview in below. 

\paragraph{BERT}
We incorporate BERT~\cite{devlin2019bert} as the backbone to generate contextualized representation of input sentence. 

\paragraph{Semantic Composition Module}
This module aims to obtain effective phrase representation guided by the contextualized representation and constituency parsing tree. To refine phrase representation based on the structural information and its constituencies, we design a two-level attention mechanism: 1) \emph{Attention to Tokens} and 2) \emph{Attention to Children}.

\paragraph{Phrase Node Prediction}
 \texttt{SentiBERT} is supervised by phrase-level sentiment labels. We use cross-entropy as the loss function for learning the sentiment predictor.

\subsection{Attention Networks for Sentiment Semantic Composition}
\label{bert}
In this section, we describe the attention networks for sentiment semantic composition in detail. 

We first introduce the notations.  $\textrm{s}=[w_1, w_2, ..., w_n]$ denotes a sentence which consists of $n$ words. $\rm{phr}=[\rm{phr}_1,\rm{phr}_2,...,\rm{phr}_m]$ denotes the phrases on the binary constituency tree of sentence $\textrm{s}$. $\mathbf{h}=[\mathbf{h}_1,\mathbf{h}_2,...,\mathbf{h}_n]$ is the contextualized representation of tokens after forwarding to a fully-connected layer, where $\mathbf{h}_t \in \mathbb{R}^{d}$. Suppose $st_i$ and $en_i$ are beginning and end indices of the $i$-th phrase where $w_{st_i},w_{st_i+1},...,w_{en_i}$ are constituent tokens of the $i$-th phrase. The corresponding token representation is $[\mathbf{h}_{st_i},\mathbf{h}_{st_i+1},...,\mathbf{h}_{en_i}]$. $\mathbf{p}_{i}$ is the phrase representation of the $i$-th phrase.
\paragraph{Attention to Tokens}
\label{att}
Given the contextualized representations of the tokens covered by a phrase. We first generate phrase representation $\mathbf{v}_i$ for a phrase $i$ by the following attention network.
\begin{equation}
\label{att_t}
\begin{aligned}
& \mathbf{q}_i=\frac{1}{en_i-st_i+1}\sum_{j=st_i}^{en_i} \mathbf{h}_j, \\
& t_j=\mathrm{Attention}(\mathbf{q}_i,\mathbf{h}_j), st_i \leq j \leq en_i, \\
& a_j=\frac{\mathrm{exp}(t_j)}{\sum_{k=st_i}^{en_i} \mathrm{exp}(t_k)}, \\
& \mathbf{o}_i=\sum_{j=st_i}^{en_i} a_j \cdot \mathbf{h}_j. \\
\end{aligned}
\end{equation}

In Eq. \eqref{att_t}, we first treat the averaged representation for each token as the query, and then allocate attention weights according to the correlation with each token. $a_j$ represents the weight distributed to the $j$-th token. We concatenate the weighted sum $\mathbf{o}_i$ and $\mathbf{q}_i$ and feed it to forward networks. Lastly, we obtain the initial representation for the phrase $\mathbf{v}_i \in \mathbb{R}^{d}$ based on the representation of constituent tokens. The detailed computation of attention mechanism is shown in Appendix \ref{calatt}.
\paragraph{Attention to Children}
\label{atc}
Furthermore, we refine phrase representations in the second layer based on constituency parsing tree and the representations obtained in the first layer.
To aggregate information based on hierarchical structure, we develop the following network. For each phrase, the attention network computes correlation with its children in the binary constituency parse tree and itself. 

Assume that the indices of child nodes of the $i$-th phrase are $lson$ and $rson$. Their representations generated from the first layer are $\mathbf{v}_{i}$,  $\mathbf{v}_{lson}$, and $\mathbf{v}_{rson}$, respectively. 
We generate the attention weights  $r_{lson}$, $r_{rson}$ and $r_{i}$ over the $i$-th phrase and its left and right children by the following.
\begin{equation}
\begin{aligned}
& c_{lson}=\mathrm{Attention}(\mathbf{v}_{i},\mathbf{v}_{lson}), \\
& c_{rson}=\mathrm{Attention}(\mathbf{v}_{i},\mathbf{v}_{rson}), \\
& c_{i}=\mathrm{Attention}(\mathbf{v}_{i},\mathbf{v}_{i}), \\
& r_{lson}, r_{rson}, r_{i}=\mathrm{Softmax}(c_{lson}, c_{rson}, c_{i}). \\
\end{aligned}
\label{ac}
\end{equation}
Then the refined representation of phrase $i$ is computed by 
\begin{equation*}
\mathbf{f}_{i}  = r_{lson} \cdot \mathbf{v}_{lson} + r_{rson} \cdot \mathbf{v}_{rson} + r_{i} \cdot \mathbf{v}_{i}.
\label{agg}
\end{equation*}
Finally, we concatenate the weighted sum $\mathbf{f}_i$ and $\mathbf{v}_i$ and feed it to forward networks with $\mathrm{SeLU}$~\cite{klambauer2017self} and $\mathrm{GeLU}$ activations~\cite{hendrycks17baseline} and layer normalization~\cite{ba2016layer}, similar to \citet{spanbert} to generate the final phrase representation $\mathbf{p}_{i} \in \mathbb{R}^{d}$. Note that when the child of $i$-th phrase is token node, the attention mechanism will attend to the representation of all the subtokens the token node covers.

\subsection{Training Objective of SentiBERT}
\label{senti_train}

Inspired by BERT, the training objective of \texttt{SentiBERT} consists of two parts: 1) Masked Language Modeling. Some texts are masked and the model learn to predict them. This objective allows the model learn to capture the contextual information as in the original BERT model. 2) Phrase Node Prediction. We further consider training the model to predict the phrase-level sentiment label based on the aforementioned phrase representations. This allows \texttt{SentiBERT} lean to capture the compositional sentiment semantics.
Similar to BERT, in the transfer learning setting, pre-trained \texttt{SentiBERT} model can be used to initialize the model parameters of a downstream model. 

\section{Experiments}
We  evaluate \texttt{SentiBERT} on the SST dataset. We then evaluate \texttt{SentiBERT} in a transfer learning setting and demonstrate that the compositional sentiment semantics learned on SST can be transferred to other related tasks.

\subsection{Experimental Settings}
\label{settings}
We evaluate how effective \texttt{SentiBERT} captures the compositional sentiment semantics on SST dataset~\cite{socher2013recursive}. 

The SST dataset has several variants. 
\begin{itemize}
\item SST-phrase is a 5-class classification task that requires to predict the sentiment of all phrases on a binary constituency tree. Different from~\citet{socher2013recursive}, we test the model only on phrases (non-terminal constituents) and ignore its performance on tokens.

\item SST-5 is a 5-class sentiment classification task that aims at predicting the sentiment of a sentence. We use it to test if \texttt{SentiBERT} learns a better sentence representation through capturing compositional sentiment semantics. 

\item Similar to SST-5, SST-2 and SST-3 are 2-class and 3-class sentiment classification tasks. However, the granularity of the sentiment classes is different.
\end{itemize}

Besides, to test the transferability of \texttt{SentiBERT}, we consider several related datasets, including Twitter Sentiment Analysis~\cite{rosenthal2017semeval}, Emotion Intensity Classification~\cite{mohammad2018semeval} and Contextual Emotion Detection (EmoContext)~\cite{chatterjee2019semeval}. Details are shown in Appendix \ref{downstream}.

We build \texttt{SentiBERT} on the HuggingFace library\footnote{\url{https://github.com/huggingface}} and initialize the model parameters using pre-trained BERT-base and RoBERTa-base models whose maximum length is 128, layer number is 12, and embedding dimension is 768. For the training on SST-phrase, the learning rate is $2\times 10^{-5}$, batch size is 32 and the number of training epochs is 3. For masking mechanism, to put emphasis on modeling sentiments, the probability of masking opinion words which can be retrieved from SentiWordNet~\cite{baccianella2010sentiwordnet} is set to 20\%, and for the other words, the probability is 15\%. For fine-tuning on downstream tasks, the learning rate is $\{1\times 10^{-5}- 1\times 10^{-4}\}$, batch size is $\{16,32\}$ and the number of training epochs is $1-5$. We use Stanford CoreNLP API~\cite{manning-EtAl:2014:P14-5} to obtain binary constituency trees for the sentences of these tasks to keep consistent with the settings on SST-phrase. Note that when fine-tuning on sentence-level sentiment and emotion classification tasks, the objective is to correctly label the root of tree, instead of targeting at the \texttt{[CLS]} token representation as in the original BERT.

\subsection{Effectiveness of SentiBERT}
We first compare the proposed attention networks (\texttt{SentiBERT} w/o BERT) with the following baseline models trained on SST-phrase corpus to evaluate the effectiveness of the architecture design: 1) Recursive NN~\cite{socher2013recursive}; 2) GCN~\cite{Kipf2016Semi}; 3) Tree-LSTM~\cite{tai-etal-improved}; 4) BiLSTM~\cite{hochreiter1997long} w/ Tree-LSTM.
To further understand the effect of using contextualized representation, we compare \texttt{SentiBERT} with the vanilla pre-trained BERT and its variants which combine the four mentioned baselines and BERT. The training settings remain the same with \texttt{SentiBERT}.
We also initialize \texttt{SentiBERT} with pre-trained parameters of RoBERTa (\texttt{SentiBERT} w/ RoBERTa) and further compare it with its variants.

As shown in Table \ref{effective}, \texttt{SentiBERT} and \texttt{SentiBERT} w/ RoBERTa substantially outperforms their corresponding variants and the networks merely built on the tree.
\begin{table}
\centering
\scalebox{0.9}{
\begin{tabular}{lcc}
\toprule
\textbf{Models}                             & \textbf{SST-phrase}  & \textbf{SST-5} \\ \midrule
Recursive NN                                             & 58.33      &  46.53       \\
GCN  &  60.89 & 49.34 \\
Tree-LSTM                                             & 61.71    &  50.07         \\   
BiLSTM w/ Tree-LSTM                                           &  61.89  &  50.45           \\
\midrule
BERT w/ Mean pooling & 64.53      &  50.68   \\
BERT w/ GCN & 65.23 & 54.56 \\
BERT w/ Tree-LSTM & 67.39  & 55.89 \\
RoBERTa w/ Mean pooling & 67.73 & 56.34 \\
\midrule
\texttt{SentiBERT} w/o BERT                                             & 61.04  &   50.31 \\
\texttt{SentiBERT}                    &  68.31     &  56.10  \\
\texttt{SentiBERT} w/ RoBERTa                      & \textbf{68.98}     &  \textbf{56.87}\\
\bottomrule
\end{tabular}
}
\caption{The averaged accuracies on SST-phrase and SST-5 tasks (\%) for 5 runs. For baselines vanilla BERT and RoBERTa, we use mean-pooling on token representation of top layer to get phrase and sentence representation.}
\label{effective}
\end{table}
Specifically, we first observe that though our attention network (\texttt{SentiBERT} w/o BERT) is simple, it is competitive with Recursive NN, GCN and Tree-LSTM. Besides, \texttt{SentiBERT} largely outperforms \texttt{SentiBERT} w/o BERT by leveraging contextualized representation. Moreover, the results manifest that \texttt{SentiBERT} and \texttt{SentiBERT} w/ RoBERTa outperform the BERT and RoBERTa, indicating the importance of incorporating syntactic guidance.

\subsection{Transferability of SentiBERT}
Though the designed models are effective, we are curious how beneficial the compositional sentiment semantics learned on SST can be transferred to other tasks. 
We compare \texttt{SentiBERT} with published models BERT, XLNet, RoBERTa and their variants on benchmarks mentioned in Section \ref{settings}. Specifically, `BERT' indicates the model trained on the raw texts of the SST dataset. `BERT w/ Mean pooling' denotes the model trained on SST, whose phrase and sentence representation is computed by mean pooling on tokens. `BERT w/ Mean pooling' merely leverages the phrases' range information rather than syntactic structural information.

\paragraph{Sentiment Classification Tasks}
The evaluation results of sentence-level sentiment classification on the three tasks are shown in Table \ref{transfer}.
\begin{table}
\centering
\scalebox{0.72}{
\begin{tabular}{lcccc}
\toprule
\textbf{Models}                             & \textbf{SST-2 (Dev)}  & \textbf{SST-3} & \textbf{Twitter} \\ \midrule
BERT &   92.39 & 73.78  & 70.0       \\
BERT w/ Mean pooling & 92.33           & 74.35               &    69.7     \\
XLNet &  93.23       & 75.89  &    70.7     \\
RoBERTa &   94.31 & 78.04 &  71.1         \\
\midrule
\texttt{SentiBERT} w/o BERT &   86.57 & 68.32 &  64.9        \\
\texttt{SentiBERT} w/o Masking &  92.48 & 76.95 &  70.7   \\
\texttt{SentiBERT} w/o Pre-training &  92.44 & 76.78 &  70.8   \\
\texttt{SentiBERT}        &  92.78       & 77.11     &  70.9 \\
\texttt{SentiBERT} w/ RoBERTa                     &  \textbf{94.72}       &  \textbf{78.69}  & \textbf{71.5}   \\
\bottomrule
\end{tabular}
}
\caption{The averaged results on sentence-level sentiment classification (\%) for 5 runs. For SST-2,3, the metric is accuracy; for Twitter Sentiment Analysis, we use averaged recall value.}
\label{transfer}
\end{table}
Despite the difference among tasks and datasets, from experimental results, we find that \texttt{SentiBERT} has competitive performance compared with various baselines. \texttt{SentiBERT} achieves higher performance than the vanilla BERT and XLNet in tasks such as SST-3 and Twitter Sentiment Analysis. Besides, \texttt{SentiBERT} significantly outperform \texttt{SentiBERT} w/o BERT. This demonstrates the importance of leveraging pre-trained BERT model. Moreover, \texttt{SentiBERT} outperforms BERT w/ Mean pooling. This indicates the importance of modeling the compositional structure of sentiment.

\paragraph{Emotion Classification Tasks}
Emotion detection is different from sentiment classification. However, these two tasks are related. The task aims to classify fine-grained emotions, such as happiness, fearness, anger, sadness, etc. It is challenging compared to sentiment analysis because of various emotion types. We fine-tune \texttt{SentiBERT} and \texttt{SentiBERT} w/ RoBERTa on Emotion Intensity Classification and EmoContext. 
\begin{table}
\centering
\scalebox{0.68}{
\begin{tabular}{lcccccc}
\toprule
\textbf{Models}                             & \textbf{Emotion Intensity} & \textbf{EmoContext} \\ \midrule
BERT  & 65.2 &  73.49          \\
RoBERTa  &  66.4 & 74.20            \\
\midrule
\texttt{SentiBERT} w/o Pre-training       & 66.0      & 73.81       \\
\texttt{SentiBERT}        & 66.5      & 74.23       \\
\texttt{SentiBERT} w/ RoBERTa        &  \textbf{67.2} & \textbf{74.67}       \\
\bottomrule
\end{tabular}
}
\caption{The averaged results on several emotion classification tasks (\%) for 5 runs. For Emotion Intensity Classification task, the metric is averaged Pearson Correlation value of the four subtasks; for EmoContext, we follow the standard metrics used in \citet{chatterjee2019semeval} and use F1 score as the evaluation metric.}
\label{emotion}
\end{table}
Table \ref{emotion} shows the results on the two emotion classification tasks. Similar to the results in sentiment classification tasks, \texttt{SentiBERT} obtains the best results, further justifying the transferability of \texttt{SentiBERT}.

\section{Analysis}
We conduct experiments on SST-phrase using BERT-base model as backbone to demonstrate the effectiveness and interpretability of the \texttt{SentiBERT} architecture in terms of semantic compositionality. We also explore potential of the model when lacking phrase-level sentiment information. In order to simplify the analysis of the change of sentiment polarity, we convert the 5-class labels to to 3-class: the classes `very negative' and `negative' are converted to be `negative'; the classes `very positive' and `positive' are converted to be `positive'; the class `neutral' remains the same. The details of statistical distribution in this part is shown in Appendix \ref{detailana}.

We consider the following baselines to evaluate the effectiveness of each component in \texttt{SentiBERT}. First we design BERT w/ Mean pooling as a base model, to demonstrate the necessity of incorporating syntactic guidance and implementing aggregation on it. Then we compare \texttt{SentiBERT} with alternative aggregation approaches, Tree-LSTM, GCN and w/o Attention to Children.
\begin{figure}[t]
\centering
  \includegraphics[width=\linewidth, trim=0 0 20 25, clip]{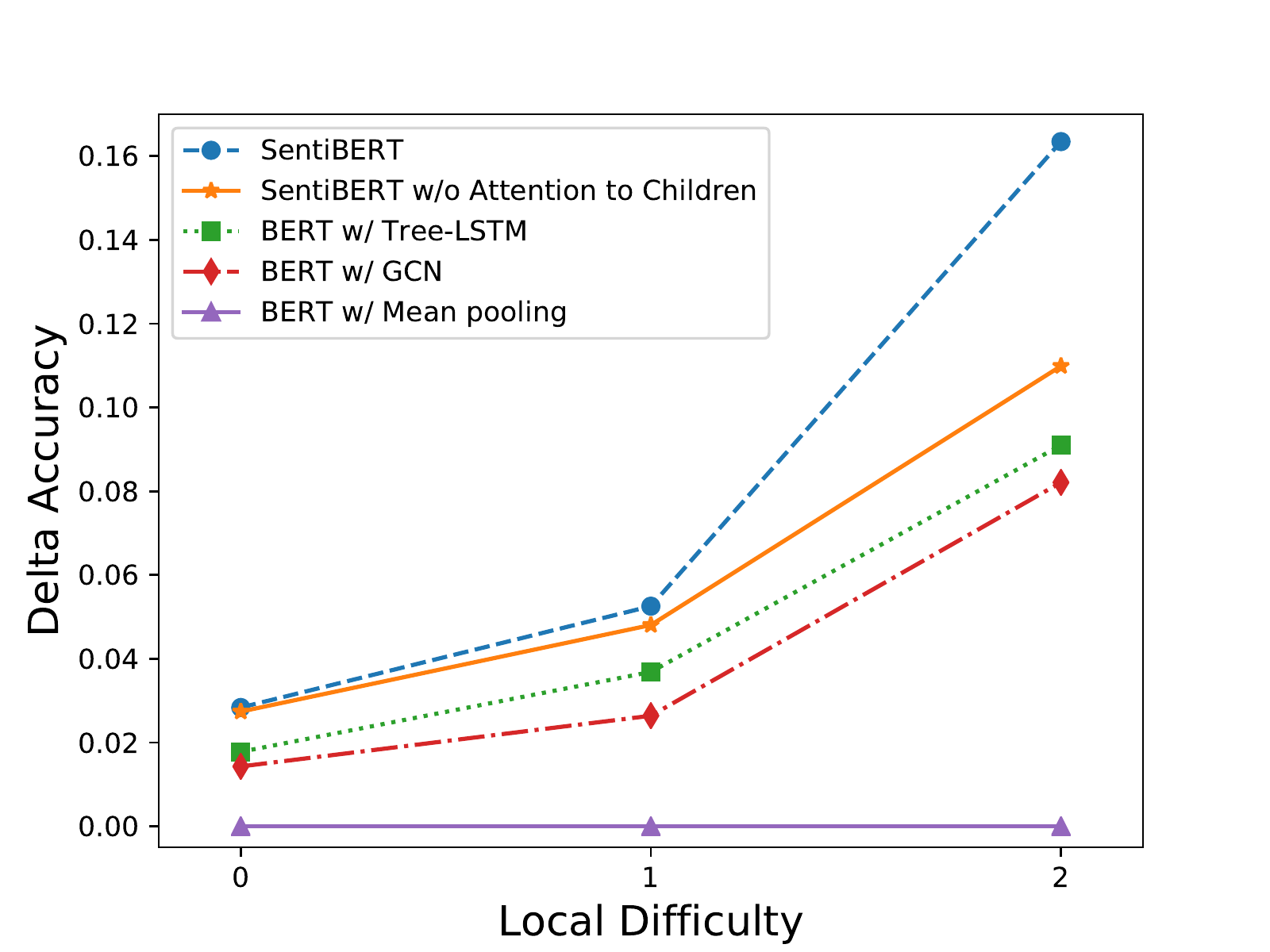}
  \caption{Evaluation for local difficulty. The figure shows the accuracy difference on phrase node sentiment prediction with BERT w/ Mean pooling for different local difficulty.}
  \label{fig:local}
\end{figure}

\subsection{Semantic Compositionality}
We investigate how effectively \texttt{SentiBERT} captures compositional sentiment semantics. We focus on how the representation in \texttt{SentiBERT} captures the sentiments when the children and parent in the constituency tree have different sentiments (i.e., sentiment switch) as shown in the red boxes of Figure \ref{intro_exp}. Here we focus on the sentiment switches between phrases. We assume that the more the sentiment switches, the harder the prediction is.

We analyze the model under the following two scenarios: \emph{local difficulty} and \emph{global difficulty}. Local difficulty is defined as the number of sentiment switches between a phrase and its children. As we consider binary constituency tree. The maximum number of sentiment switches for each phrase is 2. Global difficulty indicates number of sentiment switches in the entire constituency tree. The maximum number of sentiment switches in the test set is 23. The former is a phrase-level analysis and the latter is sentence level.

\begin{figure}[t]
\centering
  \includegraphics[width=\linewidth, trim=0 0 20 25, clip]{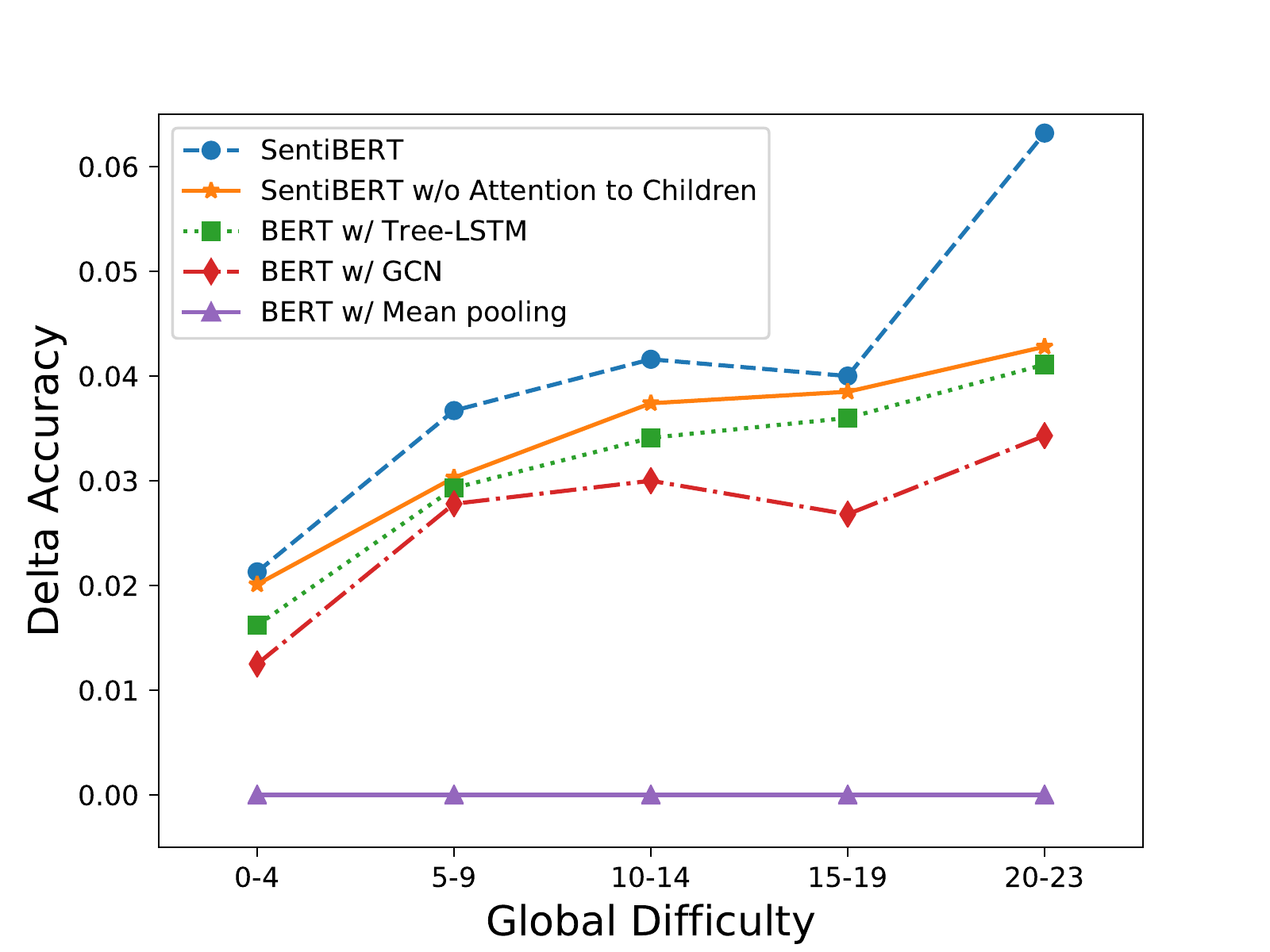}
  \caption{Evaluation for global difficulty. The figure shows the accuracy difference on phrase node sentiment prediction with BERT w/ Mean pooling for different global difficulty.}
  \label{fig:global}
\end{figure}

We compare \texttt{SentiBERT} with aforementioned baselines. We group all the nodes and sentences in the test set by local and global difficulty. Results are shown in Figure \ref{fig:local} and Figure \ref{fig:global}. Our model achieves better performance than baselines in all situations. Also, we find that with the increase of difficulty, the gap between our models and baselines becomes larger. Especially, when the sentiment labels of both children are different from the parent node (i.e., local difficulty is 2), the performance gap between \texttt{SentiBERT} and BERT w/ Tree-LSTM is about 7\% accuracy. It also outperforms the baseline BERT model with mean pooling by 15\%. This validates the necessity of structural information for semantic composition and the effectiveness of our designed attention networks for leveraging the hierarchical structures.

\subsection{Negation and Contrastive Relation}
Next, we investigate how \texttt{SentiBERT} deals with negations and contrastive relation.

\paragraph{Negation:} Since the negation words such as \textit{`no'}, \textit{`n't'} and \textit{`not'} will cause the sentiment switches, the number of negation words also reflects the difficulty of understanding sentence and its constituencies. We first group the sentences by the number of negation words, and then calculate the accuracy of the prediction on their constituencies respectively. In test set, as there are at most 
six negation words and the amount of sentences with above three negation words is small, we separate all the data into three groups.
\begin{figure}[t]
\centering
  \includegraphics[width=\linewidth, trim=0 0 20 20, clip]{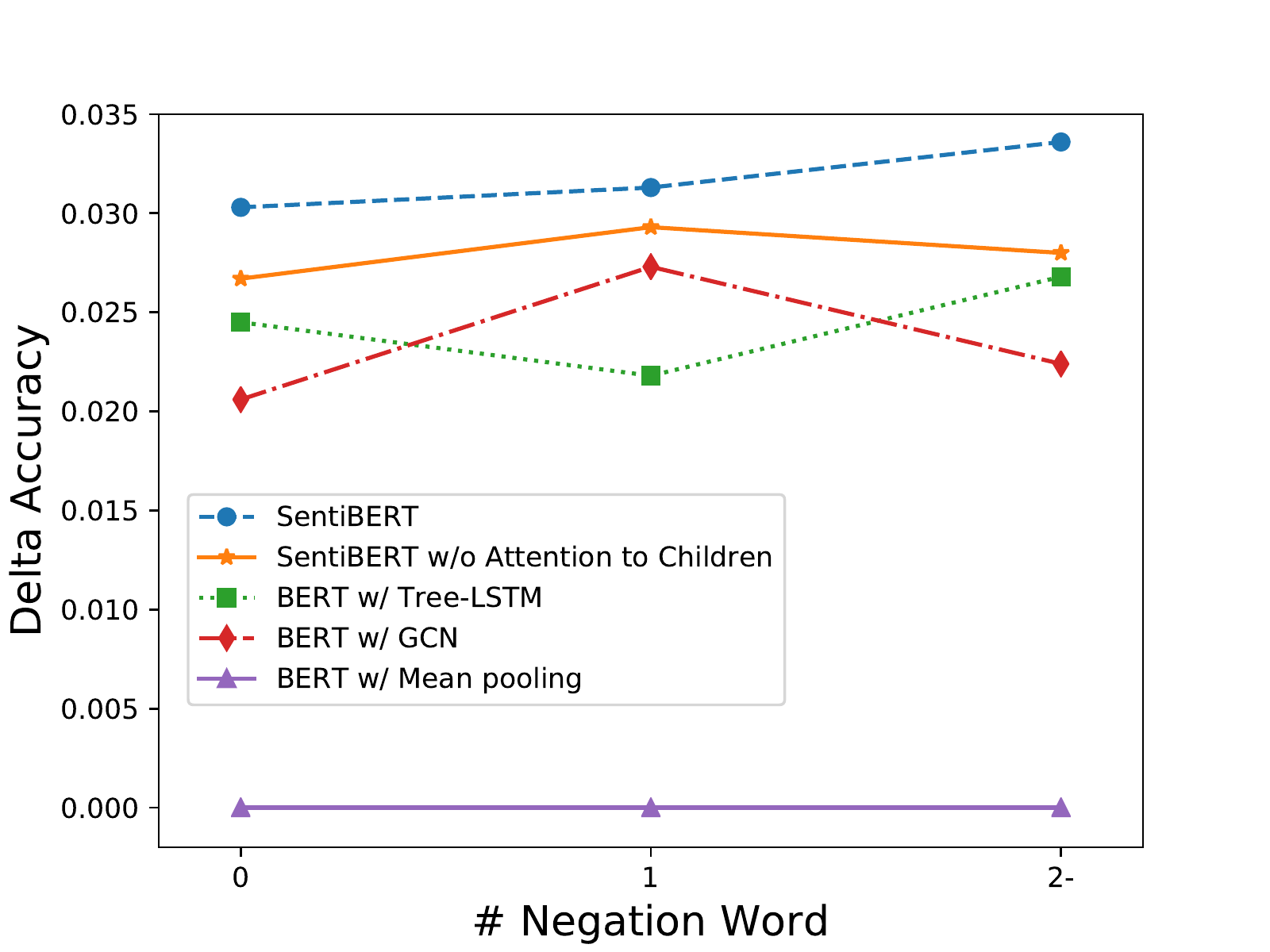}
  \caption{Evaluation for negation. We show the accuracy difference with BERT w/ Mean pooling.}
  \label{fig:neg}
\end{figure}

\begin{figure*}
\centering
\subfigure[SST-5]{
\begin{minipage}{0.3\textwidth}
\centering
\vspace{-1mm}
  \includegraphics[width=\textwidth, trim=0 0 20 22, clip]{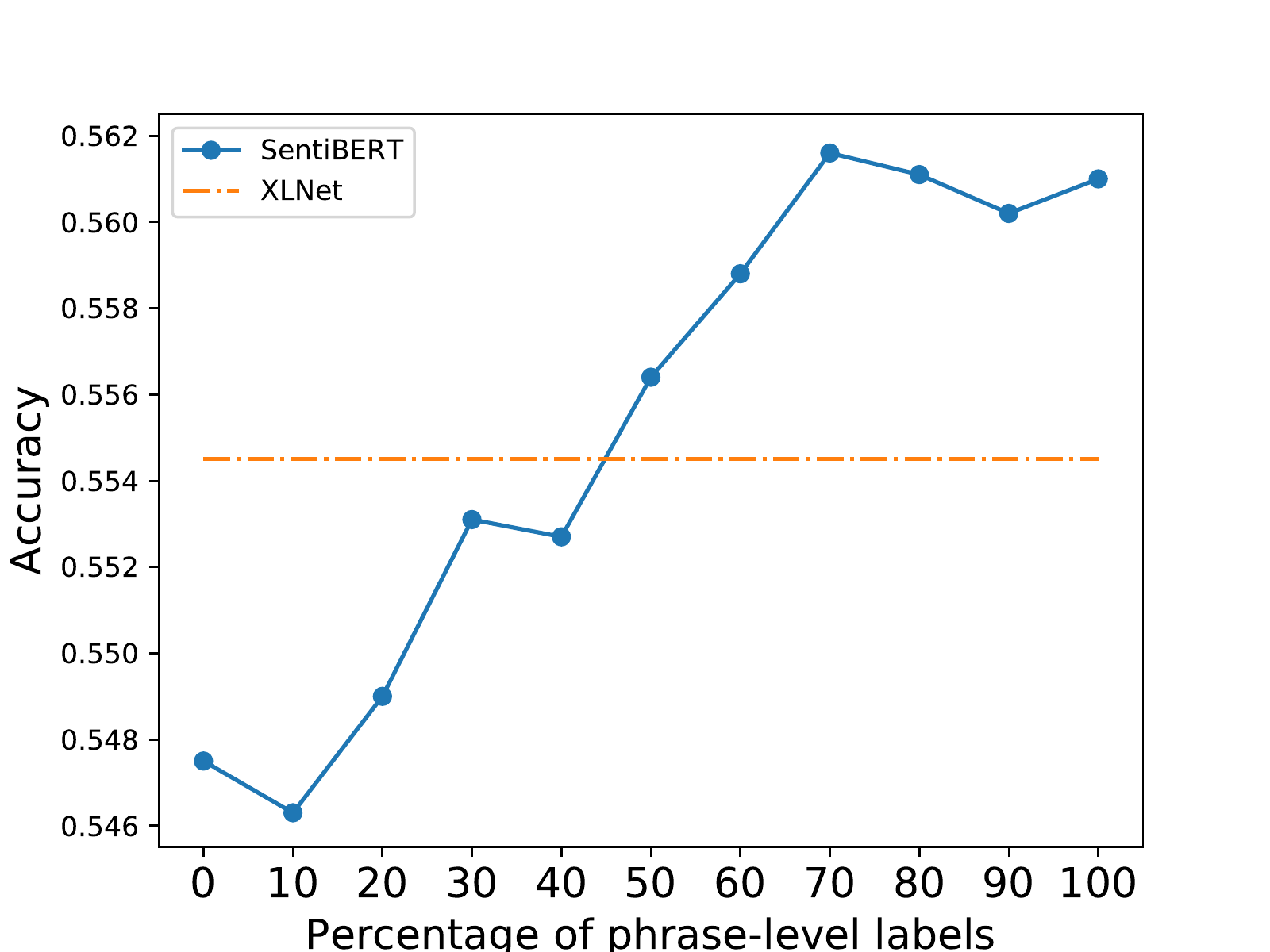}
  \vspace{-1mm}
\end{minipage}
}
\subfigure[SST-3]{
\begin{minipage}{0.3\textwidth}
\centering
\vspace{-1mm}
  \includegraphics[width=\textwidth, trim=0 0 20 22, clip]{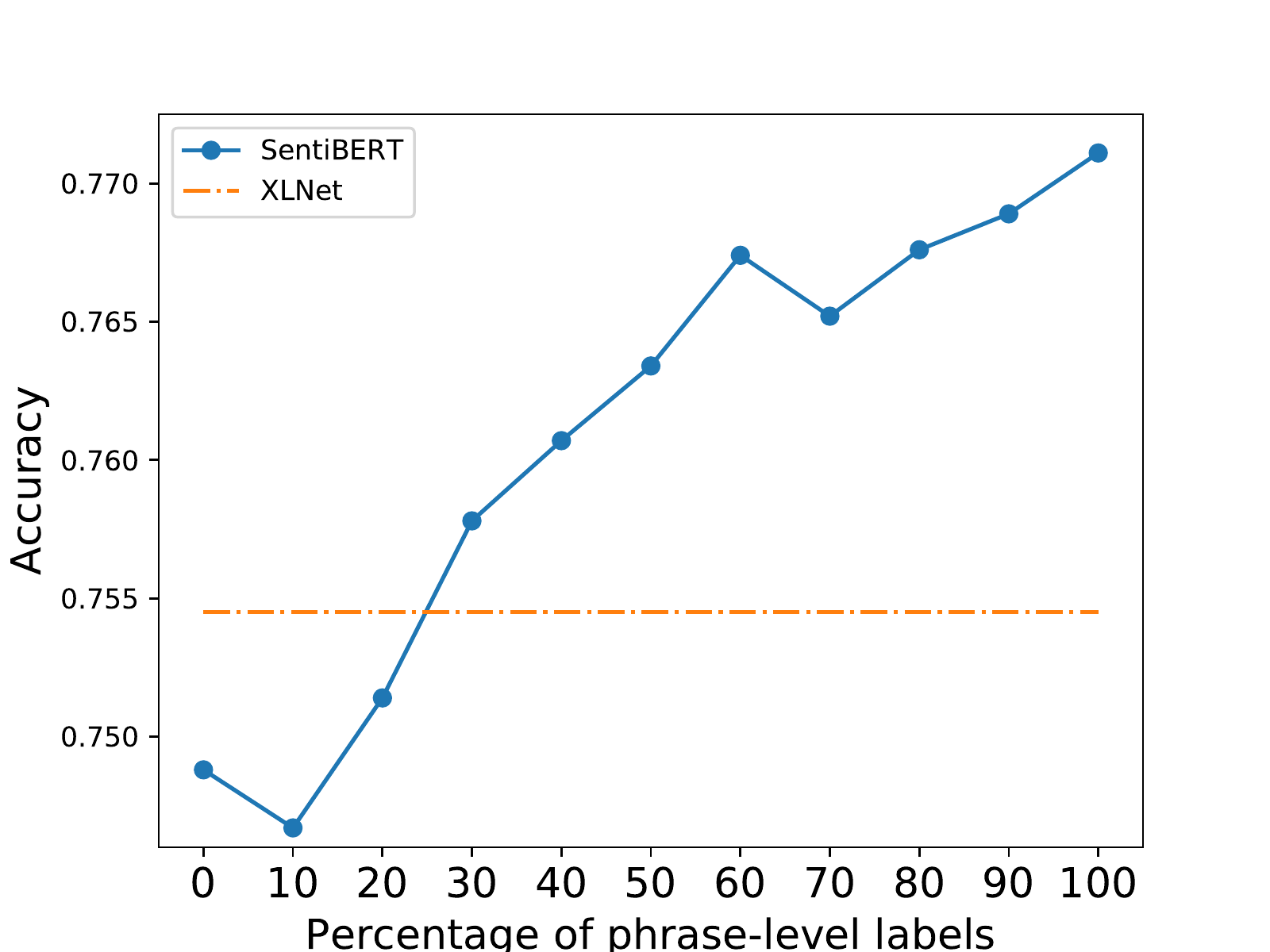}
  \vspace{-1mm}
\end{minipage}
}
\subfigure[Twitter Sentiment Analysis]{
\begin{minipage}{0.3\textwidth}
\centering
\vspace{-1mm}
  \includegraphics[width=\textwidth, trim=0 0 20 22, clip]{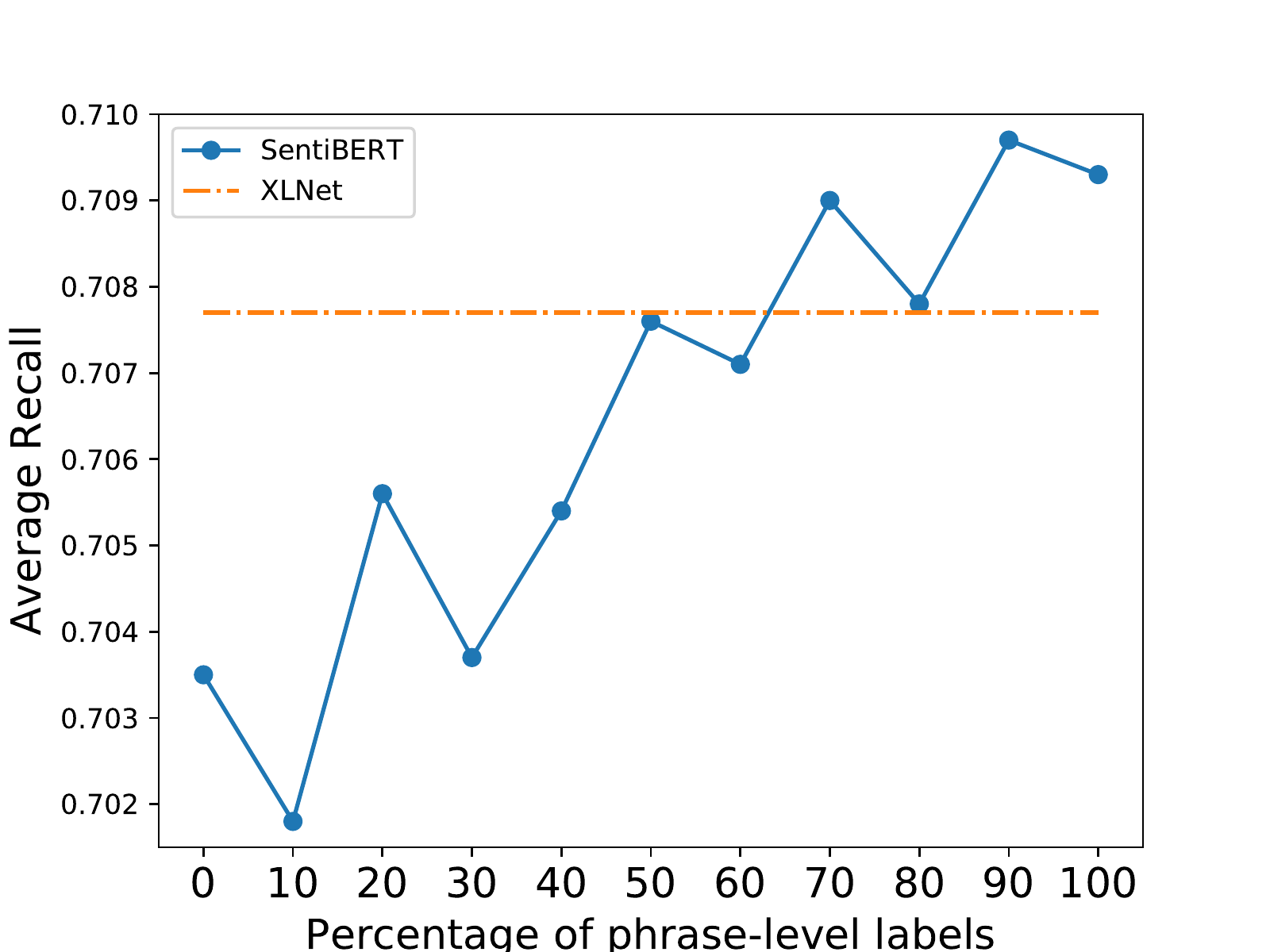}
  \vspace{-1mm}
\end{minipage}
}
\vspace{-3mm}
\caption{The results of \texttt{SentiBERT} trained with part of the phrase-level labels on SST-5, SST-3 and Twitter Sentiment Analysis. We show the averaged results of 5 runs.}
\label{fig:gamma}
\end{figure*}

Results are provided in Figure \ref{fig:neg}. We observe \texttt{SentiBERT} performs the best among all the models. Similar to the trend in local and global difficulty experiments, the gap between \texttt{SentiBERT} and other baselines becomes larger with increase of negation words. The results show the ability of \texttt{SentiBERT} when dealing with negations.
\begin{table}
\centering
\scalebox{0.82}{
\begin{tabular}{lc}
\toprule
\textbf{Models}                                  & \textbf{Accuracy} \\
\midrule
BERT w/ Mean pooling                                             &   26.1          \\
\midrule
BERT w/ Tree-LSTM                                             &    28.5         \\
BERT w/ GCN                                             &   29.4          \\
\texttt{SentiBERT} w/o Attention to Children                                             &    29.8         \\
\midrule
\texttt{SentiBERT}                      & \textbf{30.7}            \\
\bottomrule
\end{tabular}
}
\caption{Evaluation for contrastive relation (\%). We show the accuracy for triple-lets (`X but Y', `X', `Y'). X and Y must be phrases in our experiments.}
\label{contra}
\end{table}

\begin{figure}[t!]
\centering
  \includegraphics[width=\linewidth, trim=0 40 0 10, clip]{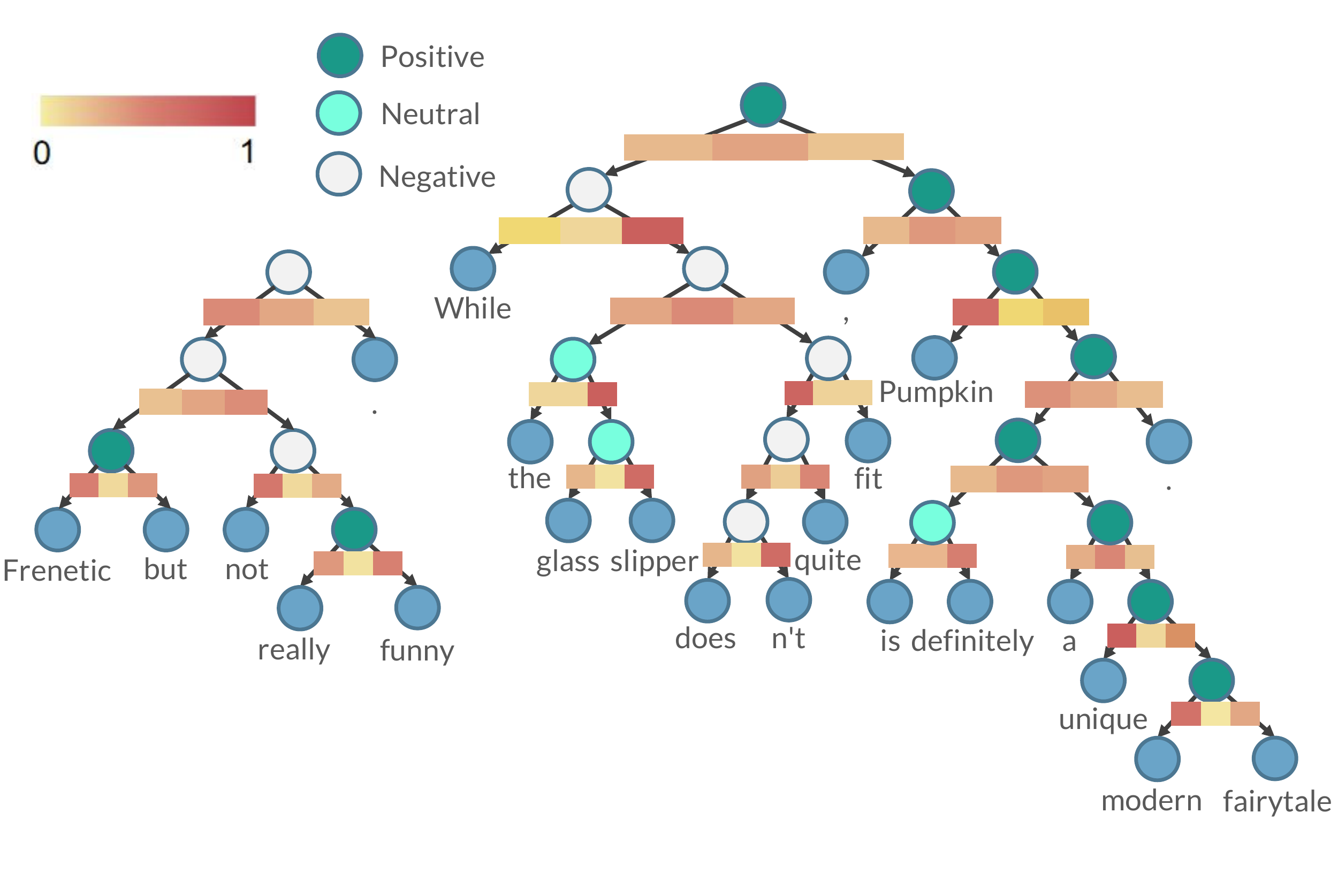}
  \caption{Cases for interpretability of compositional sentiment semantics. The three color blocks between parents and children are the attention weights distributed to left child, the phrase itself and right child.}
  \label{fig:vis}
\end{figure}

\paragraph{Contrastive Relation:} We evaluate the effectiveness of \texttt{SentiBERT} with regards to tackling contrastive relation problem. Here, we focus on the contrastive conjunction \textit{``but''}. We pick up the sentences containing word `but' of which the sentiments of left and right parts are different. In our analysis, a `X but Y' can be counted as correct if and only if the sentiments of all the phrases in triple-let (`X but Y', `X' and `Y') are predicted correctly. Table \ref{contra} demonstrates the results. \texttt{SentiBERT} outperforms other variants of BERT about 1\%, demonstrating its ability in capturing contrastive relation in sentences.

\subsection{Case Study}
We showcase several examples to demonstrate how \texttt{SentiBERT} performs sentiment semantic composition. We observe the attention distribution among hierarchical structures. In Figure \ref{fig:vis}, we demonstrate two sentences of which the sentiments of all the phrases are predicted correctly. We also visualize the attention weights distributed to the child nodes and the phrases themselves to see which part might contribute more to the sentiment of those phrases.

\texttt{SentiBERT} performs well in several aspects. First, \texttt{SentiBERT} tends to attend to adjectives such as \textit{`frenetic'} and \textit{`funny'}, which contribute to the phrases' sentiment. Secondly, facing negation words, \texttt{SentiBERT} considers them and a switch can be observed between the phrases with and without negation word (e.g., \textit{`not really funny'} and \textit{`really funny'}). Moreover, \texttt{SentiBERT} can correctly analyze the sentences expressing different sentiments in different parts. For the first case, the model concentrates more on the part after \textit{`but'}.

\subsection{Amount of Phrase-level Supervision}
We are also interested in analyzing how much phrase-level supervision \texttt{SentiBERT} needs in order to capture the semantic compositionality. We vary the amount of phrase-level annotations used in training \texttt{SentiBERT}. Before training, we randomly sample 0\% to 100\% with a step of 10\% of labels from SST training set. After pre-training on them, we fine-tune \texttt{SentiBERT} on tasks SST-5, SST-3 and Twitter Sentiment Analysis. During fine-tuning, for the tasks which use phrase-level annotation, such as SST-5 and SST-3, we use the same phrase-level annotation during pre-training and the sentence-level annotation; for the tasks which do not have phrase-level annotation, we merely use the sentence-level annotation. 

Results in Figure \ref{fig:gamma} show that with about 30\%-50\% of the phrase labels on SST-5 and SST-3, the model is able to achieve competitive results compared with XLNet. Even without any phrase-level supervision, using 70\%-80\% of phrase labels in pre-training allows \texttt{SentiBERT} competitive with XLNet on the Twitter Sentiment Analysis dataset.

Furthermore, we find the confidence of about 40-50\% of phrase nodes in SST-3 task is above 0.9 and the accuracy of predicting these phrases is above 90\% on the SST dataset. Considering the previous results, we speculate if we produce part of the phrase labels on generic texts, choose the predicted labels with high confidence and add them to the original SST training set during the training process, the results might be further improved.

\section{Conclusion}
We proposed \texttt{SentiBERT}, an architecture designed for capturing better compositional sentiment semantics. \texttt{SentiBERT} considers the necessity of contextual information and explicit syntactic guidelines for modeling semantic composition. Experiments show the effectiveness and transferability of \texttt{SentiBERT}. Further analysis demonstrates its interpretability and potential with less supervision. For future work, we will extend \texttt{SentiBERT} to other applications involving phrase-level annotations.

\section*{Acknowledgement}
We would like to thank the anonymous reviewers for the helpful discussions and suggestions. Also, we would thank Liunian Harold Li, Xiao Liu, Wasi Ahmad and all the members of UCLA NLP Lab for advice about experiments and writing.  This material is based upon work supported  in  part  by a gift grant from Taboola. 

\bibliography{acl2020}
\bibliographystyle{acl_natbib}

\clearpage

\appendix

\section{Appendix}

\subsection{Details of Correlation Computation in Attention Networks}
\label{calatt}
For vectors $\mathbf{a}$ and $\mathbf{b}$, the correlation between them is computed as below:
\begin{equation}
\label{att_token}
\begin{aligned}
\mathrm{Attention}(\mathbf{a}, \mathbf{b})= & \mathrm{tanh}(\frac{1}{\alpha} \mathrm{SeLU}((W_1 \times \mathbf{a})^T \times W_3 \\ 
& \times \mathrm{SeLU}(W_2 \times \mathbf{b}))), \\
\end{aligned}
\end{equation}
where $\mathrm{SeLU}$ \cite{klambauer2017self} is an activation function and $\alpha$ equals 4. The two layers of attention networks do not share the parameters.

\subsection{Details of Downstream Tasks}
\label{downstream}
We adopt the following tasks for evaluation of sentence-level sentiment classifications:
\paragraph{SST-2,3~\cite{socher2013recursive}}
These tasks all share with the text of the SST dataset and are single-sentence sentiment classification task, of which the numbers behind indicate the number of classes. Since two of five classes in SST-5 correspond to positive and another two indicate negative, with additional neutral ones, the dataset is separated into three groups in SST-3 task. We convert the 5-class phrase-level labels in SST-5 into three classes and leverage them in the training of SST-3 task.

\paragraph{Twitter Sentiment Analysis~\cite{rosenthal2017semeval}}
For Twitter Sentiment Analysis, given a tweet, model needs to decide which sentiment it expresses: positive, negative or neutral.

\paragraph{Emotion Intensity Ordinal Classification~\cite{mohammad2018semeval}}
The task is, given a tweet and an emotion, categorizing the tweet into one of four classes of intensity that best represents tweeter's mental state. For Emotion Intensity Classification task, the metric is averaged Pearson Correlation value of the four subtasks, `happiness', `sadness', `anger' and `fearness'.

\paragraph{Emotions in Textual Conversations~\cite{chatterjee2019semeval}}
In a dialogue, given a sentence with two turns of conversation, the models needs to classify the emotion expressed in the last sentence. For EmoContext, we follow the standard metrics used in \citet{chatterjee2019semeval} and use F1 score on the three classes `happy', `sad' and `angry', except `others' class, as the evaluation metric.

The statistics of datasets is shown in Table \ref{table:dataset}.

\begin{table}[]
\centering
\scalebox{0.9}{
\begin{tabular}{lcc}
\toprule
\textbf{Dataset}                 & \textbf{Data Split}  & \textbf{\# of Classes}      \\ \midrule
SST-phrase              & 8379~/~2184   & 5                  \\
SST-2                   & 66475 / 859        & 2                  \\
SST-3                   & 8379 / 2184   & 3                  \\
SST-5                   & 8379 / 2184   & 5                  \\
Twitter                 & 50284 / 12273 & 3                  \\
EmoContext              & 30141 / 2754  & 3                  \\
\multirow{4}{*}{EmoInt} & sad: 1533 / 975       & \multirow{4}{*}{4} \\
                        & angry: 1701 / 1001     &                    \\
                        & fear: 2252 / 986      &                    \\
                        & joy: 1616 / 1105       &                     \\
\bottomrule
\end{tabular}
}
\caption{Statistics of benchmarks.}
\label{table:dataset}
\end{table}

\begin{table}[]
\centering
\begin{tabular}{lccc}
\toprule
\textbf{Local Difficulty} & \textbf{0} & \textbf{1} & \textbf{2} \\
\midrule
\textbf{Number}     &  28136 & 10174  &  1342 \\
\bottomrule
\end{tabular}
\caption{The distribution of nodes in terms of local difficulty.}
\label{locals}
\end{table}

\begin{table}[]
\centering
\scalebox{0.8}{
\begin{tabular}{lccccc}
\toprule
\textbf{Global Difficulty} & \textbf{0-4} & \textbf{5-9} & \textbf{10-14} & \textbf{15-19} & \textbf{20-23} \\
\midrule
\textbf{Number}      & 930    &  861   &  326     &   59    &   8    \\
\bottomrule
\end{tabular}
}
\caption{The distribution of nodes in terms of global difficulty.}
\label{globals}
\end{table}

\begin{table}[]
\centering
\begin{tabular}{lccc}
\toprule
\textbf{\# of Negation Words} & \textbf{0} & \textbf{1} & \textbf{2-} \\
\midrule
\textbf{Number}         & 1825  & 325  & 34 \\
\bottomrule
\end{tabular}
\caption{The distribution of nodes in terms of negation words.}
\label{negw}
\end{table}

\subsection{Details of Analysis Part}
\label{detailana}
The distribution of nodes and sentences in terms of local difficulty, global difficulty and negation words is shown in Table \ref{locals}, \ref{globals} and \ref{negw}, respectively.

\subsection{Incorporating Token Node Prediction}
\label{tokenpre}
Since the SST dataset also provides token-level sentiment labels, we combine the token node prediction with phrase node prediction learning objective together to model compositional sentiment semantics.

\begin{table}
\scalebox{0.75}{
\begin{tabular}{lcc}
\toprule
\textbf{Models}              & \textbf{SST-phrase} & \textbf{SST-5} \\
\midrule
\texttt{SentiBERT} w/ token           & 68.23               & 56.02          \\
\texttt{SentiBERT} w/ token and RoBERTa & 68.78               & 56.91          \\
\midrule
\texttt{SentiBERT}                    & 68.31               & 56.10          \\
\texttt{SentiBERT} w/ RoBERTa          & \textbf{68.98}               & \textbf{56.87}   \\
\bottomrule
\end{tabular}
}
\caption{The results after incorporating token node prediction. `Token' denotes token node prediction.}
\label{token}
\end{table}

Results are shown in Table \ref{token}. We observe that the results drops a bit after additionally incorporating token-level sentiment information. This may be because the phrase sentiment is composed but the token sentiment mainly depends on the meaning of the lexicon itself rather than a kind of compositional sentiment semantics. The inconsistency of the training objectives may result in the performance drop.

\end{document}